\documentclass[twoside,11pt]{article}

\usepackage{jmlr2e}
\usepackage{cancel}
\usepackage{subcaption}
\usepackage{amsmath}
\usepackage{bbm}
\usepackage{caption}

\newcommand{\bx}{\mathbf{x}}

\newcommand{\mbb}{\mathbb}
\newcommand{\mcl}{\mathcal}
\newcommand{\bu}{\mathbf{u}}
\newcommand{\by}{\mathbf{y}}

\newcommand{\hbu}{\hat{\mathbf{u}}}

\newcommand{\lp}{\left(}
\newcommand{\rp}{\right)}
\newcommand{\mm}{\mathbf{m}}

\DeclareMathOperator*{\argmin}{arg\,min}
\DeclareMathOperator*{\argmax}{arg\,max}

\ShortHeadings{ICML 2020 Workshop on Real World Experiment Design and Active Learning}{Workshop on Real World Experiment Design and Active Learning}
\firstpageno{1}

\begin{document}

\title{Efficient Graph-Based Active Learning with Probit Likelihood via Gaussian Approximations}

\author{\name Kevin Miller \email millerk22@math.ucla.edu \\ 
\name Hao Li \email lihao0809@math.ucla.edu \\
\name Andrea L. Bertozzi \email bertozzi@math.ucla.edu \\
       \addr Department of Mathematics\\
       University of California, Los Angeles\\
       Los Angeles, CA 90025, USA
    }

\editor{}

\maketitle

\begin{abstract}
We present a novel adaptation of active learning to graph-based semi-supervised learning (SSL) under non-Gaussian Bayesian models. 
We present an approximation of non-Gaussian distributions to adapt previously Gaussian-based acquisition functions to these more general cases. We develop an efficient rank-one update for applying ``look-ahead'' based methods as well as model retraining. We also introduce a novel ``model change'' acquisition function based on these approximations that further expands the available collection of active learning acquisition functions for such methods. 
\end{abstract}


\section{Introduction}
Active learning in semi-supervised learning (SSL) seeks to alleviate the issue of trying to train machine learning classifiers with few labeled data but ubiquitous unlabeled data. While there are a few different formulations of active learning, we focus on the {\it pool-based} active learning paradigm as opposed to online or streaming-based active learning~\citep{settles_active_2012}. That is, the active learner has access to a fixed ``pool'' of unlabeled data points from which it can decide the next training point. We consider querying only a single point at a time, as opposed to {\it batch-mode} active learning~\citep{hoi_semi-supervised_2008}. Let $Z = \{1,2,\ldots, N\}$ index a set of input feature vectors $X = \{\bx_i\}_{i=1}^N$ among which $\mcl L \subset Z$ have known labels $\{y_j\}_{j \in \mcl L}$. We assume the {\it binary classification} case, in which the labels reside in $y_j \in \{\pm 1\}$ (or $\{0, 1\}$). In pool-based active learning, most methods alternate between: (1) training a model given the current labeled data $\mcl L, \{y_j\}_{j \in \mcl L}$ and (2) choosing an active learning query point $k^\ast$ in the unlabeled set $\mcl U = Z - \mcl L$ according to an {\it acquisition function}. 
We can classify most methods into a few categories: uncertainty~\citep{settles_active_2012, houlsby_bayesian_2011, gal_deep_2017}, margin~\citep{tong_support_2001, balcan_agnostic_2006, jiang_minimum-margin_2019}, clustering~\citep{dasgupta_hierarchical_2008, maggioni_learning_2019}, and look-ahead~\citep{zhu_combining_2003, cai_maximizing_2013} acquisition functions. 
Active learning methods have been proposed for graph-based SSL models which use a similarity graph to represent the geometric relationships between points in the dataset, such as Gaussian Random Field (GRF) models ~\citep{zhu_semi-supervised_2003,bertozzi_diffuse_2016, bertozzi_uncertainty_2018}. Active learners implementing look-ahead expected risk~\citep{zhu_combining_2003, jun_graph-based_2016}, 
model posterior covariance~\citep{ji_variance_2012,ma_sigma_2013}, and other measures of uncertainty~\citep{kushnir_diffusion-based_2020} have been produced for the GRF model of~\citep{zhu_semi-supervised_2003}. The conditional distribution of this foundational GRF model is a harmonic function on the graph and hence is referred to as the {\it Harmonic Functions} (HF) model.

Our contributions are (1) provide a unifying framework for active learning in many graph-based SSL models, (2) introduce an adaptation of non-Gaussian Bayesian models to allow for efficient calculations previously done only on Gaussian models, and (3) introduce a novel ``model change'' active learning acquisition function built around our adaptation.

\section{Graph-Based SSL Models}

Consider input data $X$ with index set $Z$; we create a similarity graph $G(Z, W)$ with edge weights $W_{ij} = \kappa(\bx_i, \bx_j) \ge 0$ calculated by a similarity kernel $\kappa$. 
In this paper we consider both the {\it unnormalized graph Laplacian matrix} $L_u = D - W$ or {\it normalized graph Laplacian matrix} $L_n = D^{-1/2}(D - W)D^{-1/2}$, where $D = \operatorname{diag}(d_1, d_2, \ldots, d_N), d_i = \sum_{j \neq i} W_{ij}$ is the diagonal {\it degree matrix}. 
As $L = L_u, L_n$ are both positive semi-definite, then with $\tau>0$, $L_\tau = \tau^{-2}(L + \tau^2 I)$ is positive definite and therefore $\mcl N(0, L_\tau^{-1})$ is a well-defined Bayesian prior distribution.
Define a real-valued function on the nodes of the graph $u : Z \rightarrow \mbb R$, $\bu \in \mbb R^N$ whose values reflect the classification of the data points.
In graph-based SSL, given the current labeled set $\mcl L$, one seeks the solution to the optimization problem
\begin{equation} \label{eq:gbssl}
    \bu^\ast = \argmin_{\bu \in \mbb R^N }  \frac{1}{2} \langle \bu, L_\tau \bu \rangle + \sum_{j \in \mcl L} \ell(u_j, y_j) =: \argmin_{\bu \in \mbb R^N } J_\ell(\bu; \by),
\end{equation}
where $\ell: \mbb R \times \mbb R \rightarrow [0, \infty)$ is a chosen loss function and $\by \in \mbb R^{|\mcl L|}$ is a vector of labels $y_j$. Common loss functions include $\ell(x,y) = (x-y)^2/2\gamma^2$ and $\ell(x,y) = -\log \Psi_\gamma(xy)$, where $\Psi_\gamma(t) = \int_{-\infty}^{t} \psi_\gamma(s) ds$ is the cumulative distribution function (CDF) of a log-concave probability density function (PDF) $\psi_\gamma(s)$. 

This variational perspective has a probabilistic counterpart, from which Bayesian statistical methods can provide useful ways for devising well-principled acquisition functions. We can view the objective function in Eqn~\ref{eq:gbssl} as the negative log of an associated Bayesian posterior distribution. 
In the case of $\ell(x,y) = (x-y)^2/2\gamma^2$, we model the likelihood of observations $\by | \bu$ by $\mcl N(P\bu, \gamma^2I_{|\mcl L|})$ where $P :\mbb R^N \rightarrow \mbb R^{|\mcl L|}$ is the projection of $\bu$ onto the labeled indices $\mcl L$. This likelihood is Gaussian and therefore the posterior $\mathbbm{P}(\bu | \by)$ is Gaussian $N(\mm, C)$ with covariance $C = \lp L_\tau^{-1} +  P^TP/\gamma^2\rp^{-1}$ and mean $\mm = C P^T \by/\gamma^2$.
We refer to this as the {\it Gaussian Regression} (GR) model. The Gaussian structure of this posterior distribution allows us to efficiently  calculate the posterior mean and covariance, including look-ahead calculations. Although the prior is Gaussian, the posterior distribution for general loss functions $\ell$  
is not necessarily Gaussian. 
The key idea behind our method is to approximate a non-Gaussian distribution with a suitable Gaussian distribution to exploit the efficient calculations of the look-ahead posterior mean and covariance. This more general formulation allows us to use more realistic models for classification than just regression. 
An example of such a non-Gaussian posterior occurs when the loss function is $\ell(x,y) = -\log \Psi_\gamma(xy)$. In this case, the likelihood is derived from the model $y_j = \mathrm{Sign}(u_j + \eta_j)$, where $\eta_j \sim \psi_\gamma$~\citep{hoffmann_consistency_2020}. We refer to this as the {\it Probit} model.


Some common acquisition functions originally derived for Gaussian models are
\begin{itemize}
    \item {\bf MBR}~\citep{zhu_combining_2003} $k_{MBR} = \argmin_{k \in \mcl U} \mathbbm{E}_{y_k | \mm} \left[\sum_{i=1}^N \operatorname{Err}(i, \mm^{k, y_k}) \right]$
    \item {\bf VOpt}~\citep{ji_variance_2012} $k_{V} = \argmax_{k \in \mcl U} \frac{1}{\gamma^2 + C_{k,k}}\|C_{:,k}\|_2^2$
    \item {\bf $\Sigma$Opt}~\citep{ma_sigma_2013} $k_{\Sigma} = \argmax_{k \in \mcl U} \frac{1}{\gamma^2 + C_{k,k}}\langle \mathbbm{1}, C_{:,k}\rangle$
\end{itemize}
where $\operatorname{Err}(i, \mm^{k, y_k})$ is the estimated risk on the $i^{th}$ data point of the look-ahead mean $\mm^{k,y_k}$. These acquisition functions are originally defined on the HF model~\citep{zhu_combining_2003}, but have been generalized here to fit the GR model. To recover the HF model's acquisition functions, let $\gamma = 0$, $y_j \in \{0,1\}$, and the posterior covariance $C$ be defined only on the unlabeled nodes per the conditional nature of the HF model.

\subsection{Laplace Approximation of the Probit Model}

{\it Laplace approximation} is a popular technique for approximating non-Gaussian distributions with a Gaussian distribution~\citep{rasmussen_gaussian_2006}. 
We approximate the Probit posterior  with the Gaussian distribution:
\begin{align}\label{eq:gauss-approx-probit}
    \hat{\mathbbm{P}}(\bu | \by) &= \mcl N (\hbu, \hat{C}), \ \hbu = \argmin_{\bu \in \mbb R^N} J_\ell(\bu ; \by), \ \hat{C} = \lp \nabla \nabla J_\ell(\bu; \by)|_{\bu = \hbu}  \rp^{-1}.
\end{align}
The mean of this Gaussian distribution $\hbu$ is the {\it maximum a posteriori} (MAP) estimator of the true Probit posterior. This Gaussian distribution is in a form in which we can apply adaptations of
acquisition functions of GR and HF models,
such as VOpt~\citep{ji_variance_2012}, $\Sigma$-Opt~\citep{ma_sigma_2013}, and MBR~\citep{zhu_combining_2003}. The Laplace approximations of the GR and HF models are indeed themselves, because the mean and MAP estimator (i.e. mode) are the same for Gaussian distributions.
Furthermore, this Laplace approximation of non-Gaussian posterior distributions incorporates labeling information that is not contained in the GR and HF models' covariance matrices.

\subsection{Look-Ahead Updates} 

Acquisition functions such as MBR need a {\it look-ahead model} with index $k$ and label $y_k$:
\[
    \argmin_{\bu \in \mbb R^N} J^{k}(\bu ; \by, y_k) := \argmin_{\bu \in \mbb R^N} \frac{1}{2} \langle \bu, L_\tau \bu \rangle + \sum_{j \in \mcl L} \ell(u_j, y_j) + \ell(u_k, y_k).
\]
This is simply the updated graph-based SSL problem, having added the index $k$ and associated label $y_k$ to the labeled data. As mentioned previously, one convenience of Gaussian models is that we can solve for the look-ahead posterior distribution's parameters from the current posterior distribution {\it without expensive model retraining}. 
This is a crucial property for computing acquisition functions like MBR \citep{zhu_combining_2003}, that consider the effects of adding an index $k$ with label $y_k$ to the labeled data. 
There is no simple, closed-form solution for computing the look-ahead MAP estimator $\hbu^{k,y_k}$ from the current $\hbu$ in the Probit model (Eqn~\ref{eq:gauss-approx-probit}) because of the loss function $-\ln \Psi_\gamma(xy)$. We approximate the look-ahead update $\tilde{\bu}^{k,y_k}$ by computing a single step of Newton's Method on the look-ahead objective $J^k(\bu; \by, y_k)$, starting with the current MAP estimator $\hbu$:
\begin{align}\label{eq:na-update}
    \tilde{\bu}^{k,y_k} &= \hbu - \lp  \nabla \nabla J^k(\hbu; \by, y_k) \rp^{-1}\lp \nabla J^k(\hbu; \by, y_k) \rp  = \hbu - \frac{F(\hat{u}_k, y_k)}{1 + \hat{C}_{k,k} F'(\hat{u}_k, y_k)} \hat{C}_{:, k},
\end{align}
where $F,F'$ are the first and second derivatives of the loss function with respect to the first argument. We call this single step of Newton's method as a {\it Newton Approximation (NA) update}. This is a simple rank-one update of the MAP estimator. The update requires storing the posterior covariance matrix $\hat{C}$; this is needed for all the aforementioned Gaussian-based acquisition functions, in this context. Due to the second-order nature of Newton's method, this NA update $\tilde{\bu}^{k,y_k}$ empirically is a good approximation of the true look-ahead MAP estimator $\hbu^{k,y_k}$. 
We also derive a {\it NA posterior covariance update} similar to the GR model:
\begin{equation}\label{eq:gauss-prob-cov}
    \hat{C}^{k,y_k} = \lp  \nabla \nabla J^k(\hbu^{k,y_k}; \by, y_k) \rp^{-1} \approx \hat{C} - \frac{F'(\tilde{u}^{k,y_k}_k, y_k)}{1 + \hat{C}_{k,k} F'(\tilde{u}^{k,y_k}_k, y_k)} \hat{C}_{:, k} \hat{C}_{:,k}^T =: \tilde{C}^{k,y_k}.
\end{equation}
With these simple NA updates, we can straightforwardly apply the Gaussian-based acquisition functions to our approximation (Eqn~\ref{eq:gauss-approx-probit}) of the Probit model. Furthermore, model retraining is approximated by using these NA updates of the MAP estimator and posterior covariance, as we demonstrate in Sec.~\ref{sec:numerics}.



\subsection{Model Change (MC) Acquisition Function}\label{sec:mc-method}

Calculating the approximate change in a model (i.e. classifier) from the addition of an index $k$ and associated label $y_k$ has been investigated previously~\citep{cai_maximizing_2013, karzand_maximin_2020}. 
Employing our NA update (Eqn~\ref{eq:na-update}), we propose a MC acquisition function for our approximated Probit model in a {\it max-min} framework:
\begin{align*}
    k_{MC-P} &= \argmax_{k \in \mcl U}\ \min_{y_k \in\{\pm 1\} } \|\hbu - \hbu^{k, y_k}\|_2 \approx \argmax_{k \in \mcl U}\ \min_{y_k \in\{\pm 1\} } \left\|\frac{F(\hat{u}_k, y_k)}{1 + \hat{C}_{k,k} F'(\hat{u}_k, y_k)} \hat{C}_{:, k} \right\|_2. 
\end{align*}
\begin{figure}[t]
\vspace{-0.4in}
    \begin{subfigure}{.32\textwidth}
      \centering
      \includegraphics[width=\linewidth]{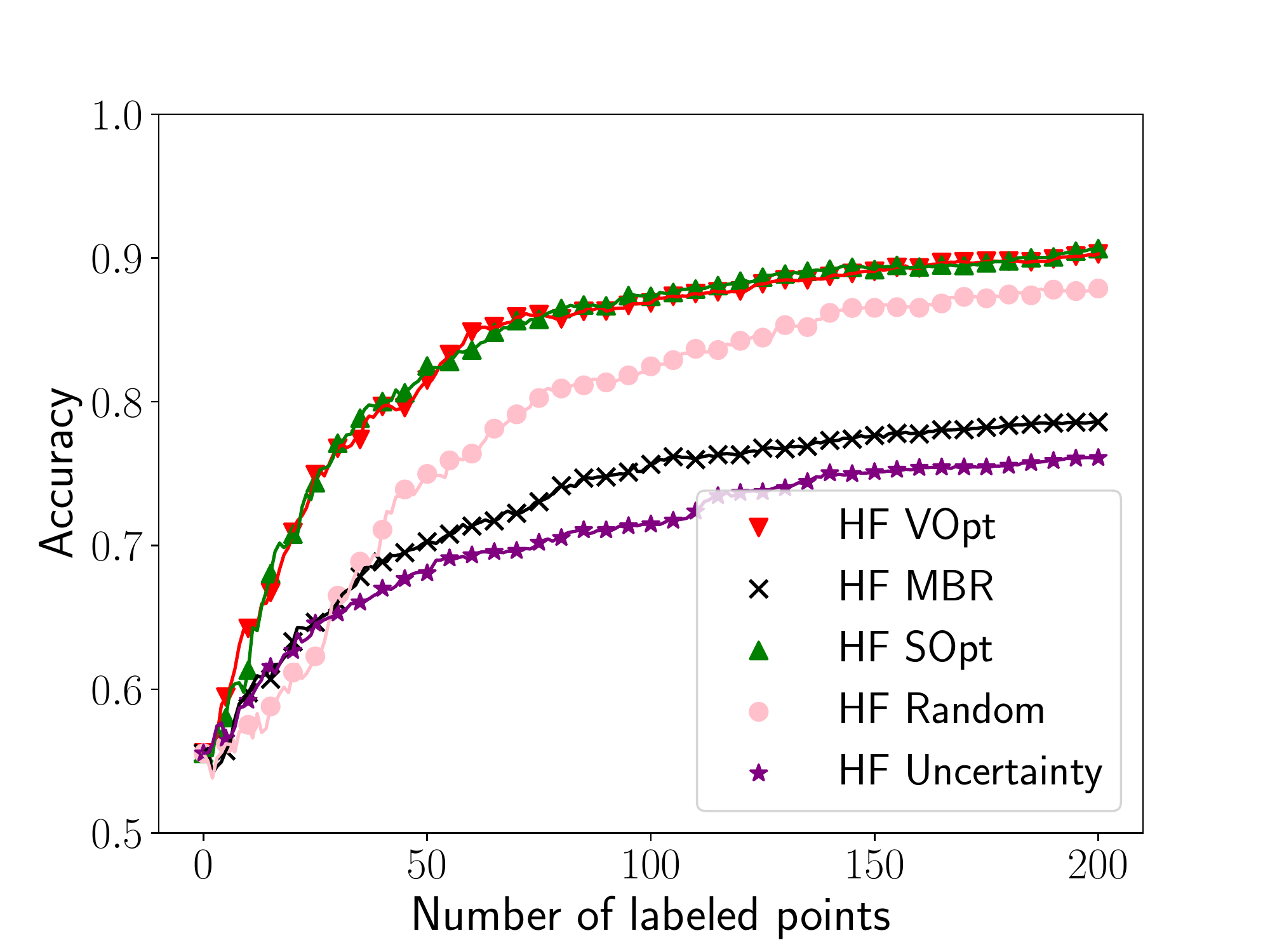}  
      \caption{HF }
      \label{fig:cb-hf}
    \end{subfigure}
    \begin{subfigure}{.32\textwidth}
      \centering
      \includegraphics[width=\linewidth]{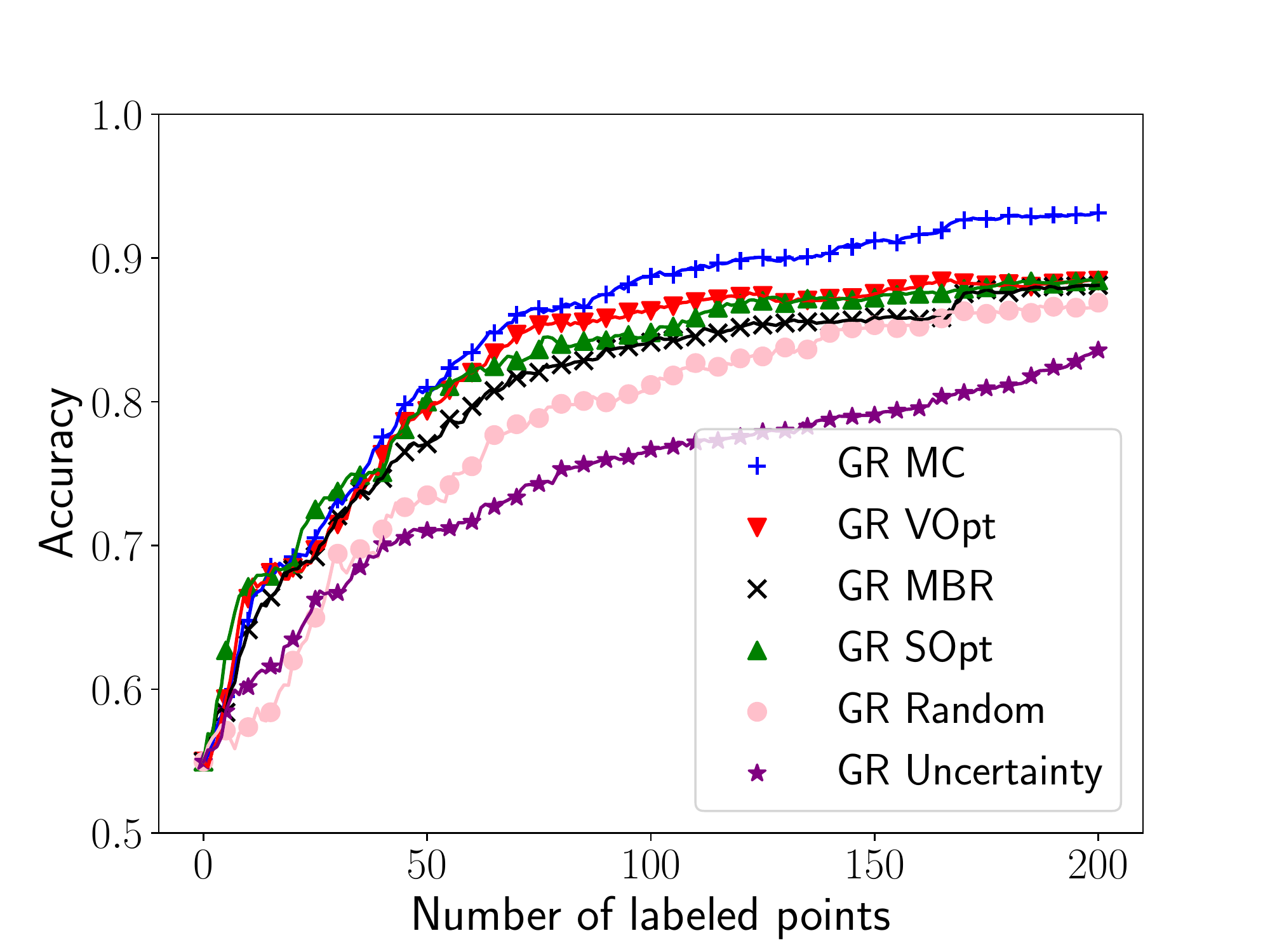}  
      \caption{GR}
      \label{fig:cb-gr}
    \end{subfigure}
    \begin{subfigure}{.32\textwidth}
      \centering
      \includegraphics[width=\linewidth]{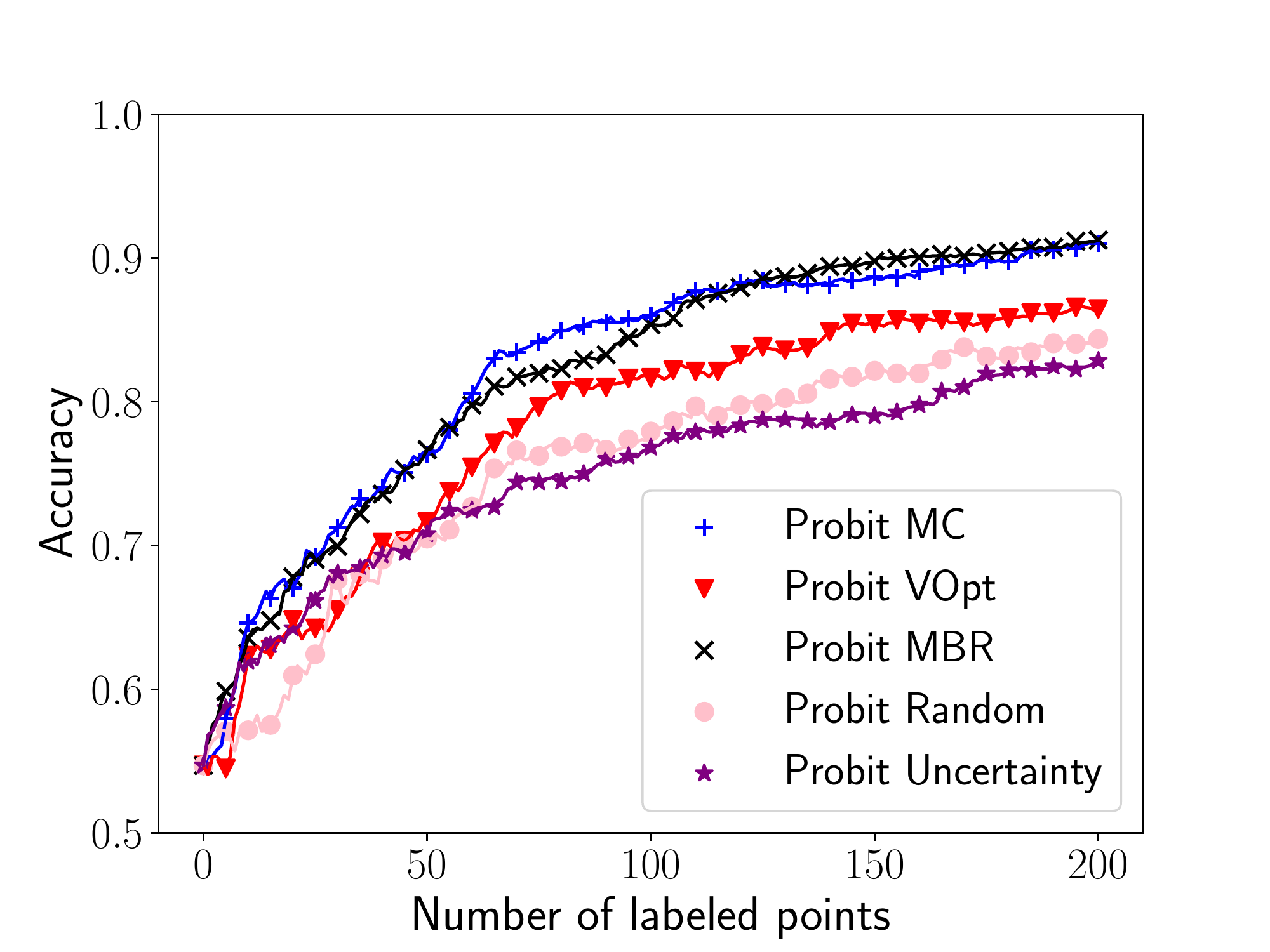}  
      \caption{Probit}
      \label{fig:cb-p2}
    \end{subfigure}
    \caption{Checkerboard Dataset Results}
    \label{fig:checkerboard}
\end{figure}


%
%
%
\vspace{-0.2in}
\section{Results}\label{sec:numerics}
\begin{figure}
\centering
    \begin{subfigure}{.32\textwidth}
        \centering        \includegraphics[width=\textwidth]{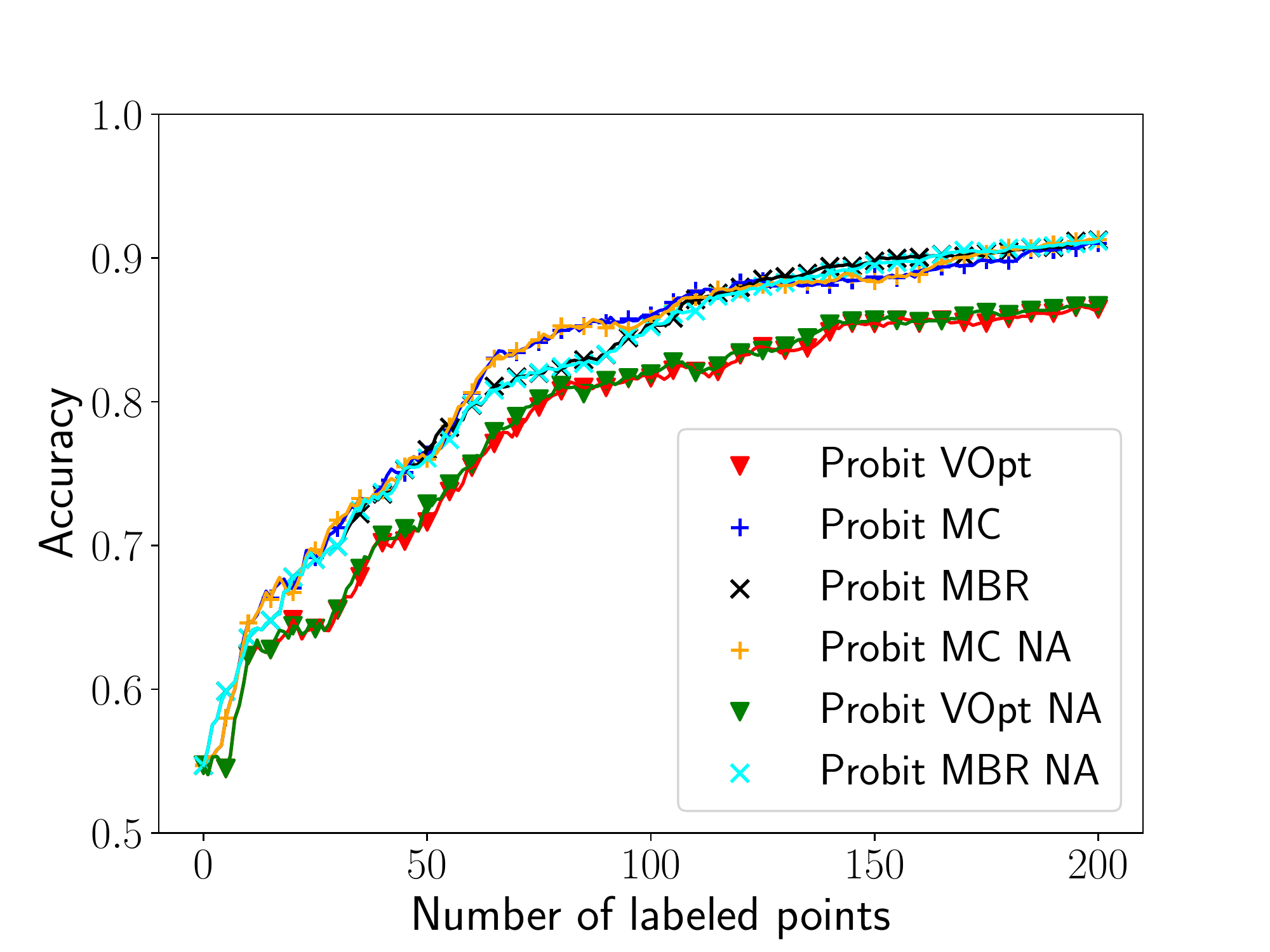}
       \caption{Checkerboard}
       \label{fig:cb-NA-close}
    \end{subfigure}
    \hskip0.09\textwidth
    \begin{subfigure}{.32\textwidth}
        \centering
        \includegraphics[width=\textwidth]{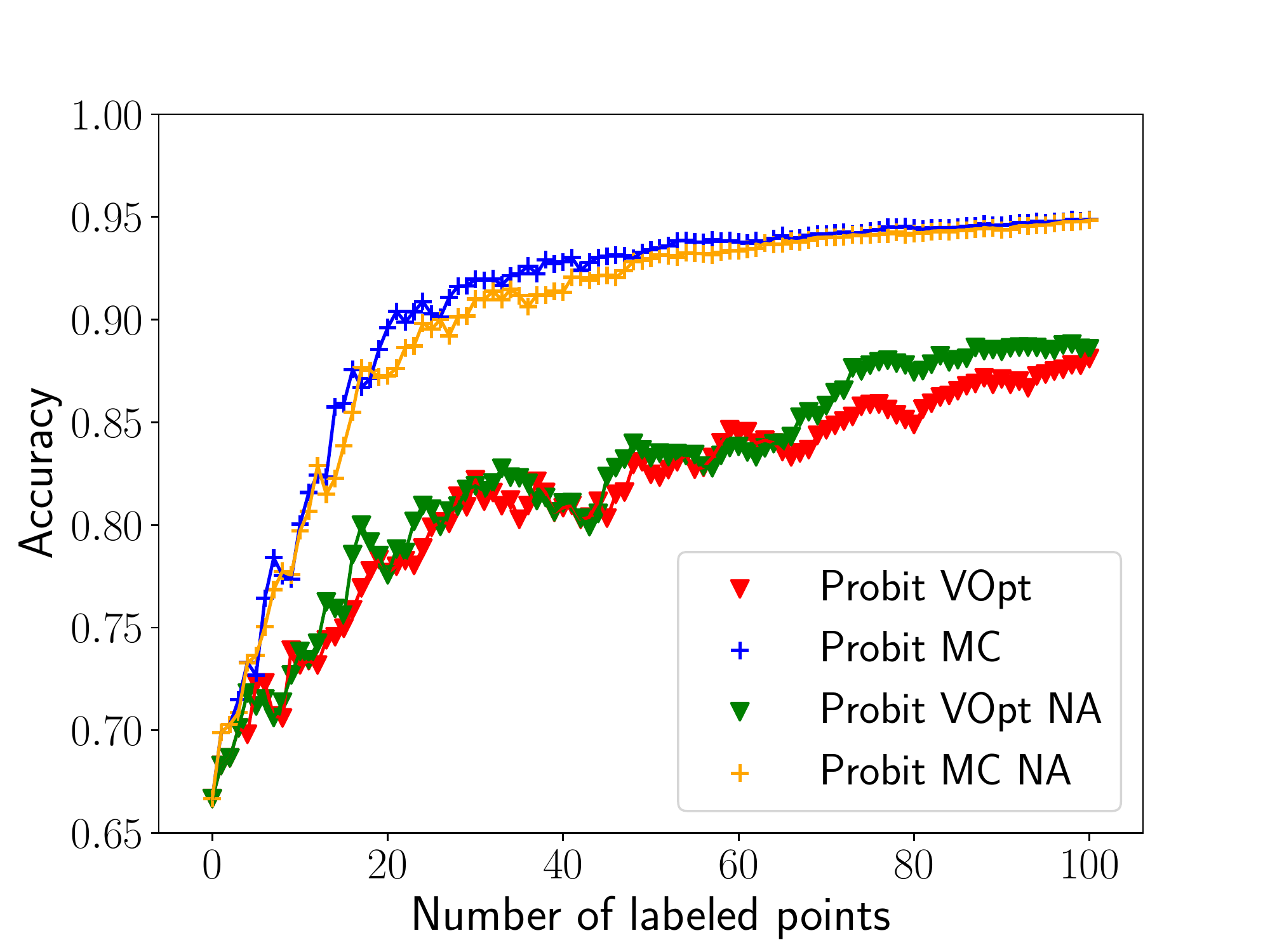}
        \caption{MNIST}
        \label{fig:mnist-NA-close}
    \end{subfigure}
    \caption{Accuracy comparison for query choices using the {\it true} posterior updates $\hbu^{k,y_k}, \hat{C}^{k,y_k}$ compared to the NA updates $\tilde{\bu}^{k,y_k}, \tilde{C}^{k,y_k}$. NA update denoted with ``NA'' in legend.}
    \label{fig:NA-close}
\end{figure}

We present numerical results demonstrating our Gaussian approximations and subsequent NA updates in the Probit model on a synthetic dataset (Checkerboard) and a real-world dataset (MNIST). In each of the {\bf HF}, {\bf GR}, and {\bf Probit} models, we show the performance of the {\bf MC} method of Sec.~\ref{sec:mc-method}, {\bf VOpt}~\citep{ji_variance_2012}, {\bf MBR}~\citep{zhu_combining_2003}, {\bf Uncertainty }\citep{settles_active_2012}, and {\bf Random}. 
We calculate the average accuracies over five trials according to the underlying SSL classifier of the acquisition function. 
After comparing accuracies across all methods with a common classifier (of the Probit model), we find that each method's query choices better improve the accuracy of its underlying classifier. In Fig.~\ref{fig:NA-close}, we demonstrate how closely the NA updates $\tilde{\bu}^{k,y_k}, \tilde{C}^{k,y_k}$ (Eqns~\ref{eq:na-update},~\ref{eq:gauss-prob-cov}) approximate the active learning choices from retraining the model (i.e. $\hbu^{k,y_k}, \hat{C}^{k,y_k}$). 
\subsection{Checkerboard Dataset}%
\begin{figure}
\centering
\vspace{-0.5in}
    \begin{subfigure}{.25\textwidth}
      \centering
      \includegraphics[width=\linewidth]{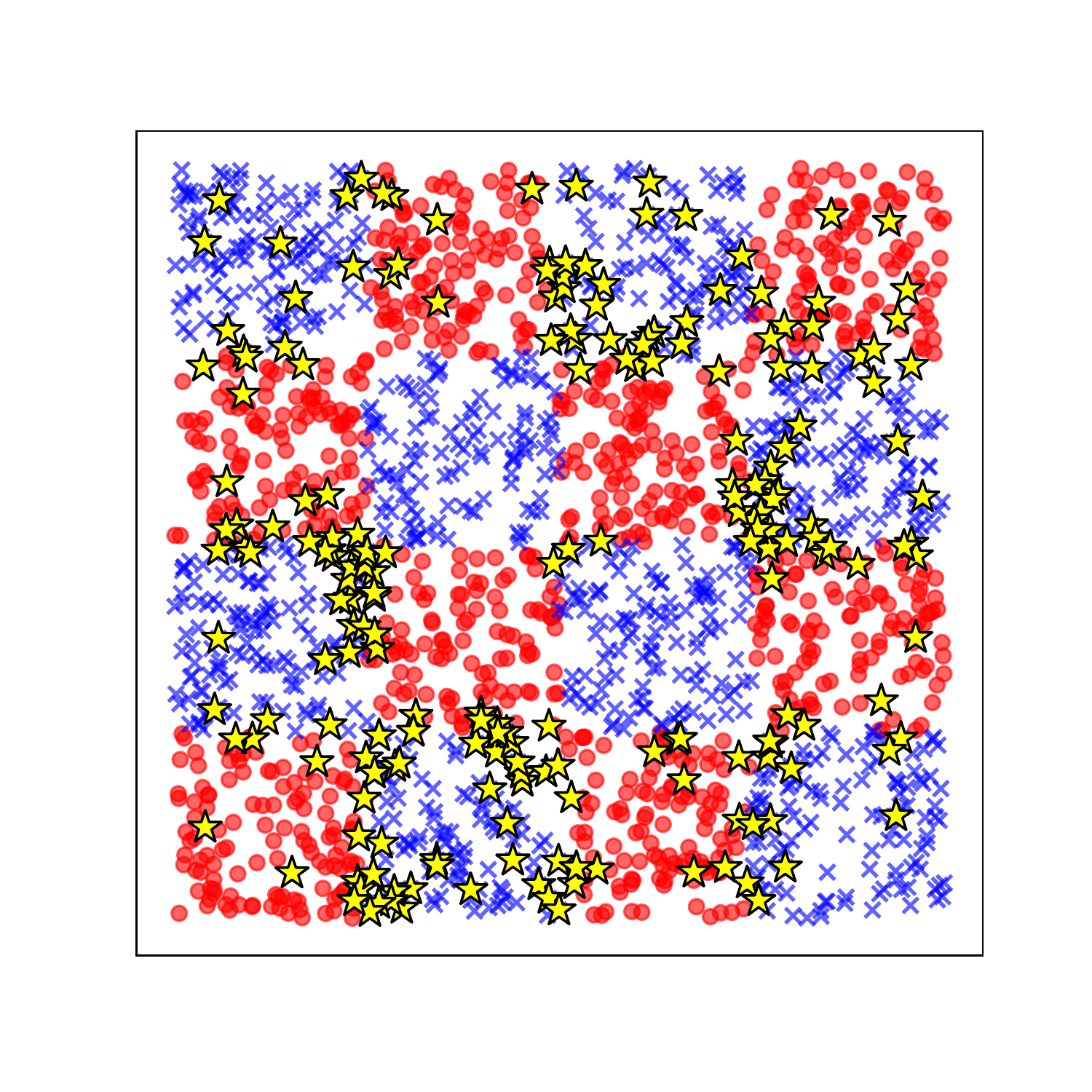} 
      \vspace{-0.4in}
      \caption{{HF-MBR}}
      \label{fig:hf-mbr-choices}
    \end{subfigure}
    \begin{subfigure}{.25\textwidth}
      \centering
      \includegraphics[width=\linewidth]{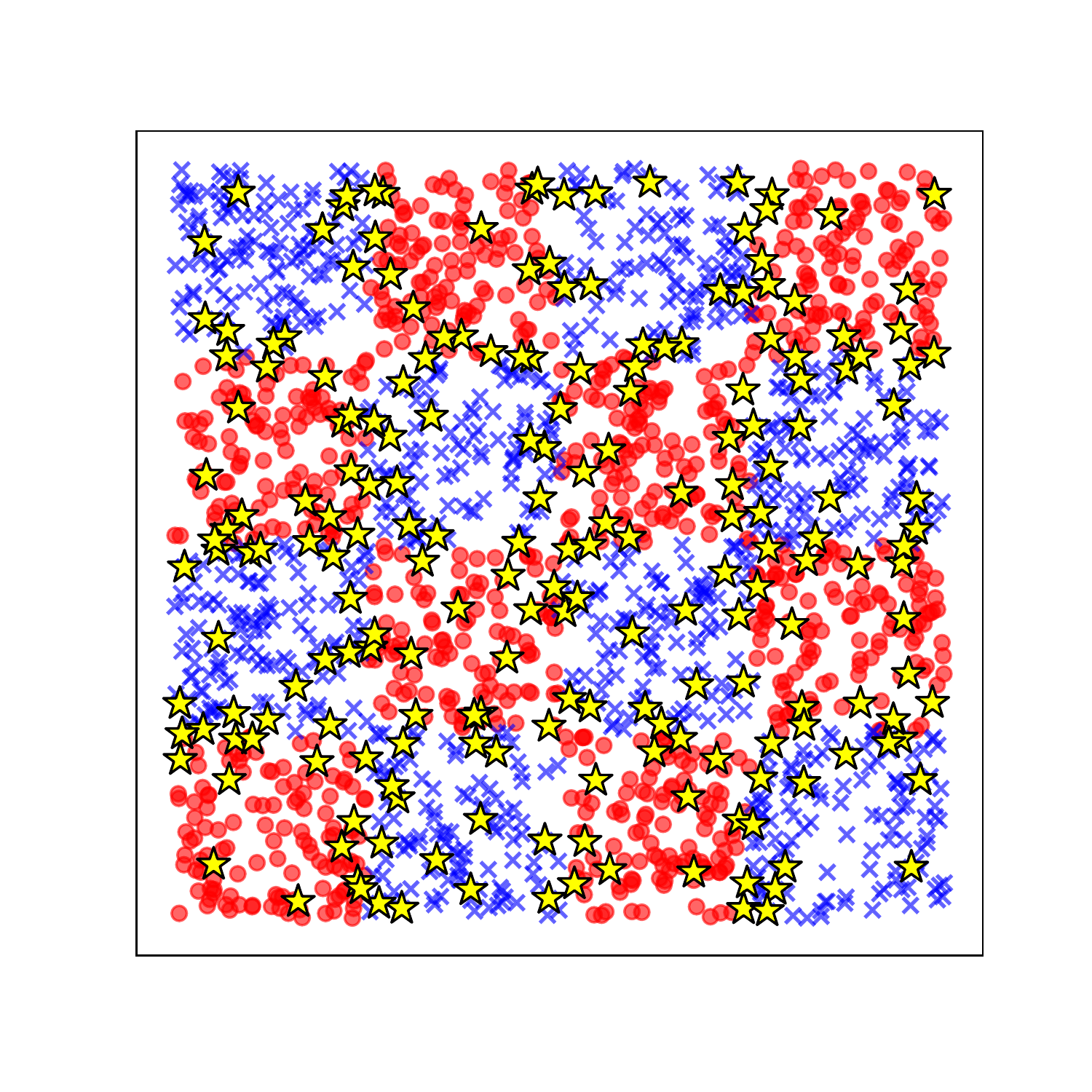}  
      \vspace{-0.4in}
      \caption{{GR-MC}}
      \label{fig:gr-mc-choices}
    \end{subfigure}
    \begin{subfigure}{.25\textwidth}
      \centering
      \includegraphics[width=\linewidth]{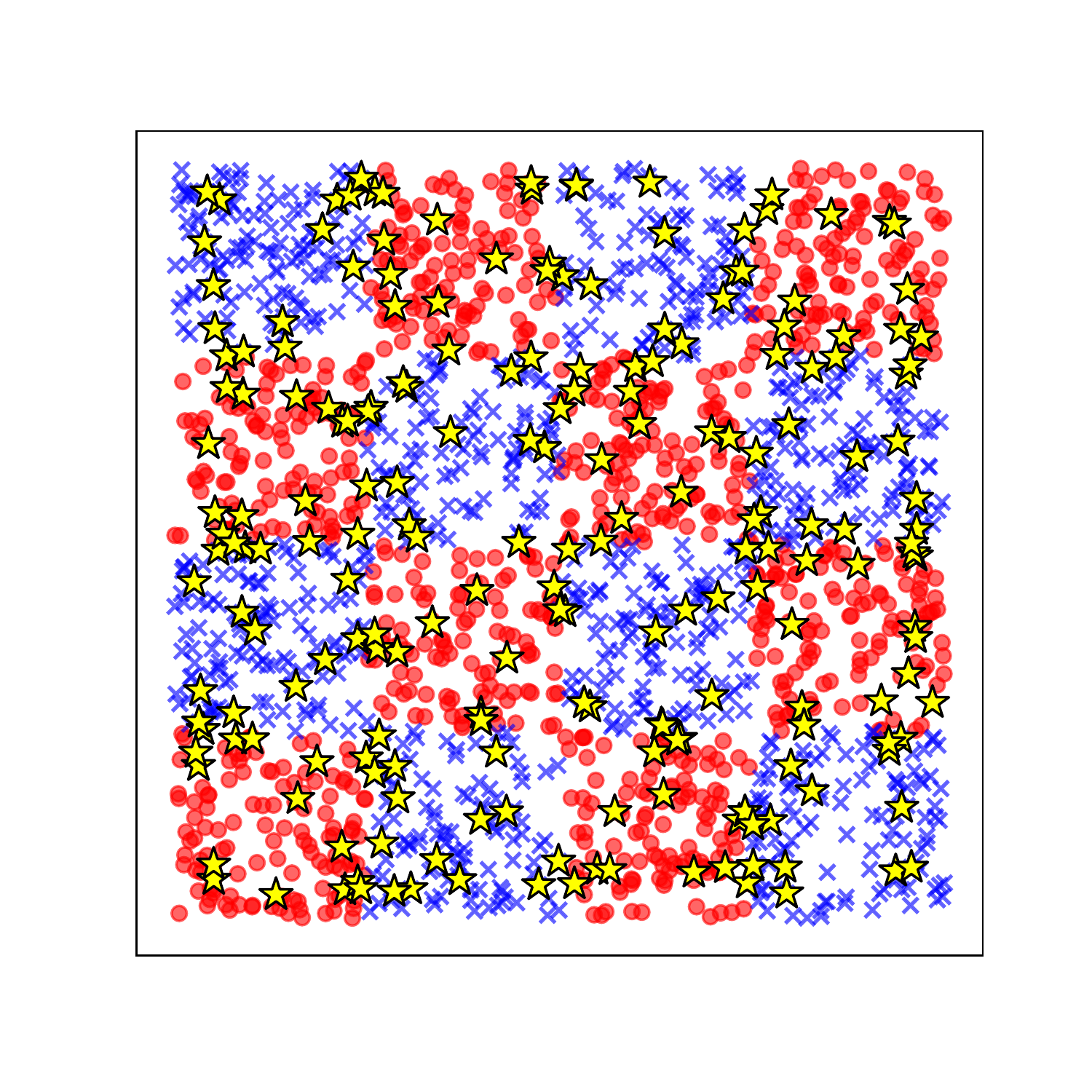}  
      \vspace{-0.4in}
      \caption{{Probit-MC}}
      \label{fig:probit-mc-choices}
    \end{subfigure}
    \vskip\baselineskip
    \vspace{-0.1in}
    \begin{subfigure}{.25\textwidth}
      \centering
      \includegraphics[width=\linewidth]{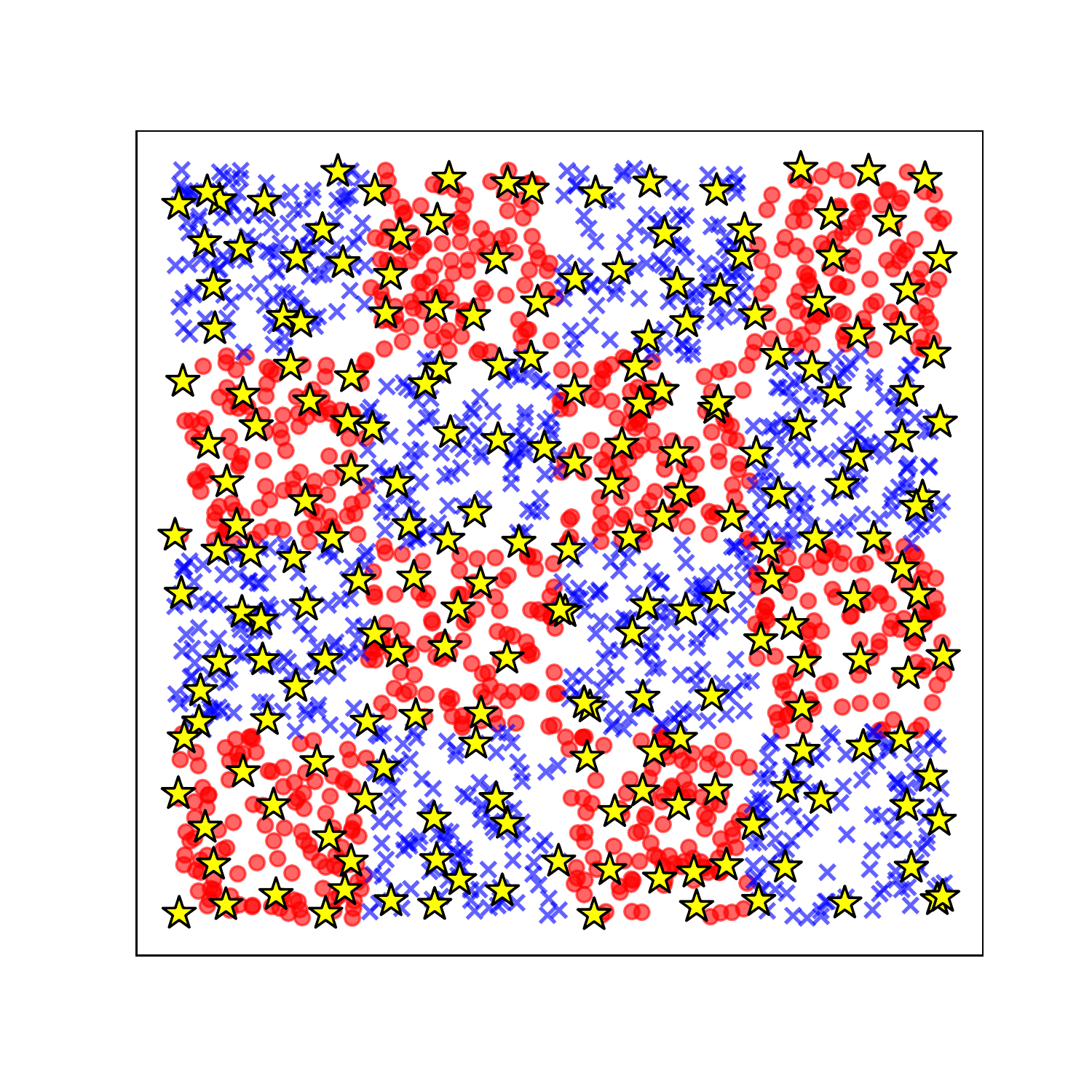} 
      \vspace{-0.4in}
      \caption{{HF-Vopt}}
      \label{fig:hf-vopt-choices}
    \end{subfigure}
    \begin{subfigure}{.25\textwidth}
      \centering
      \includegraphics[width=\linewidth]{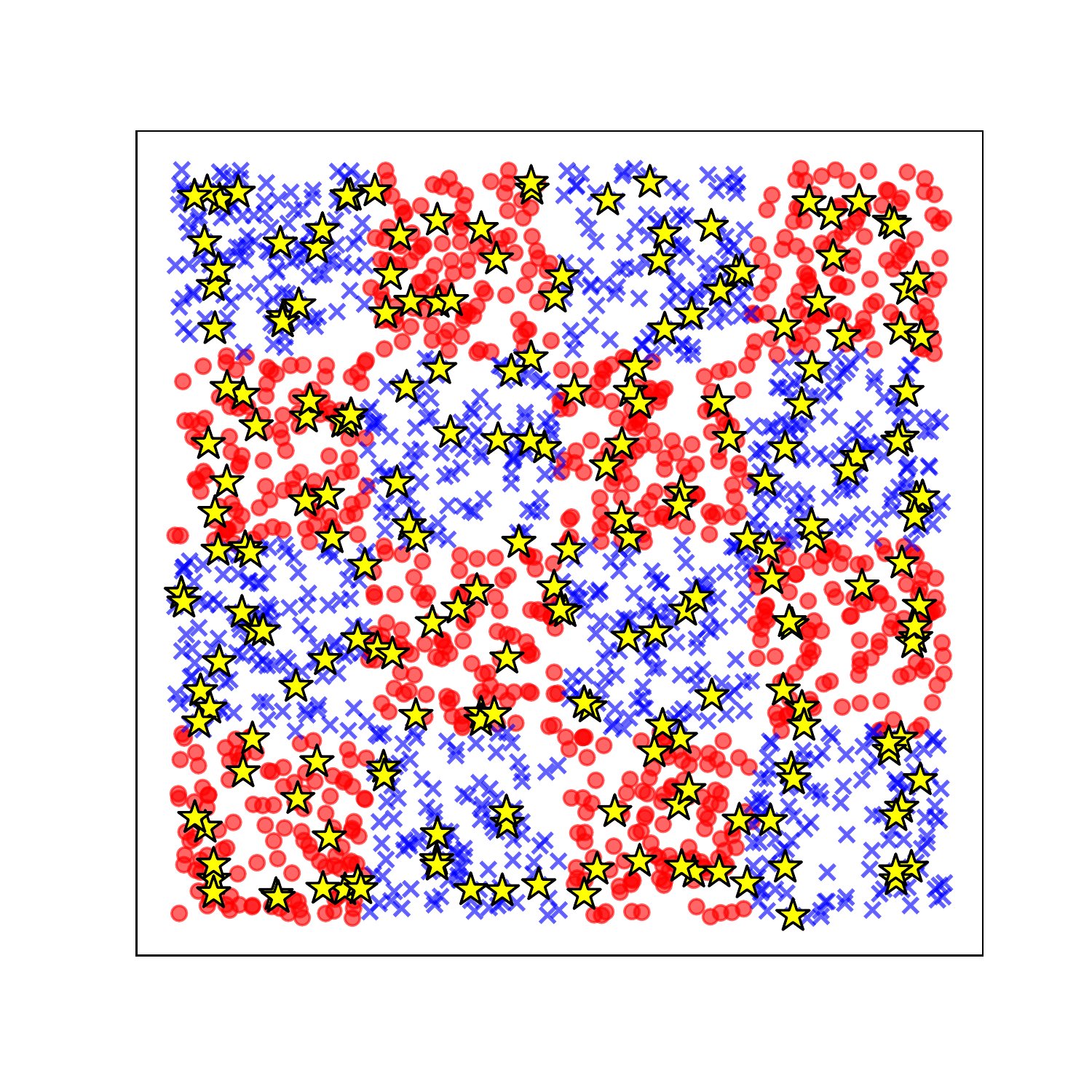}  
      \vspace{-0.4in}
      \caption{{GR-VOpt}}
      \label{fig:gr-vopt-choices}
    \end{subfigure}
    \begin{subfigure}{.25\textwidth}
      \centering
      \includegraphics[width=\linewidth]{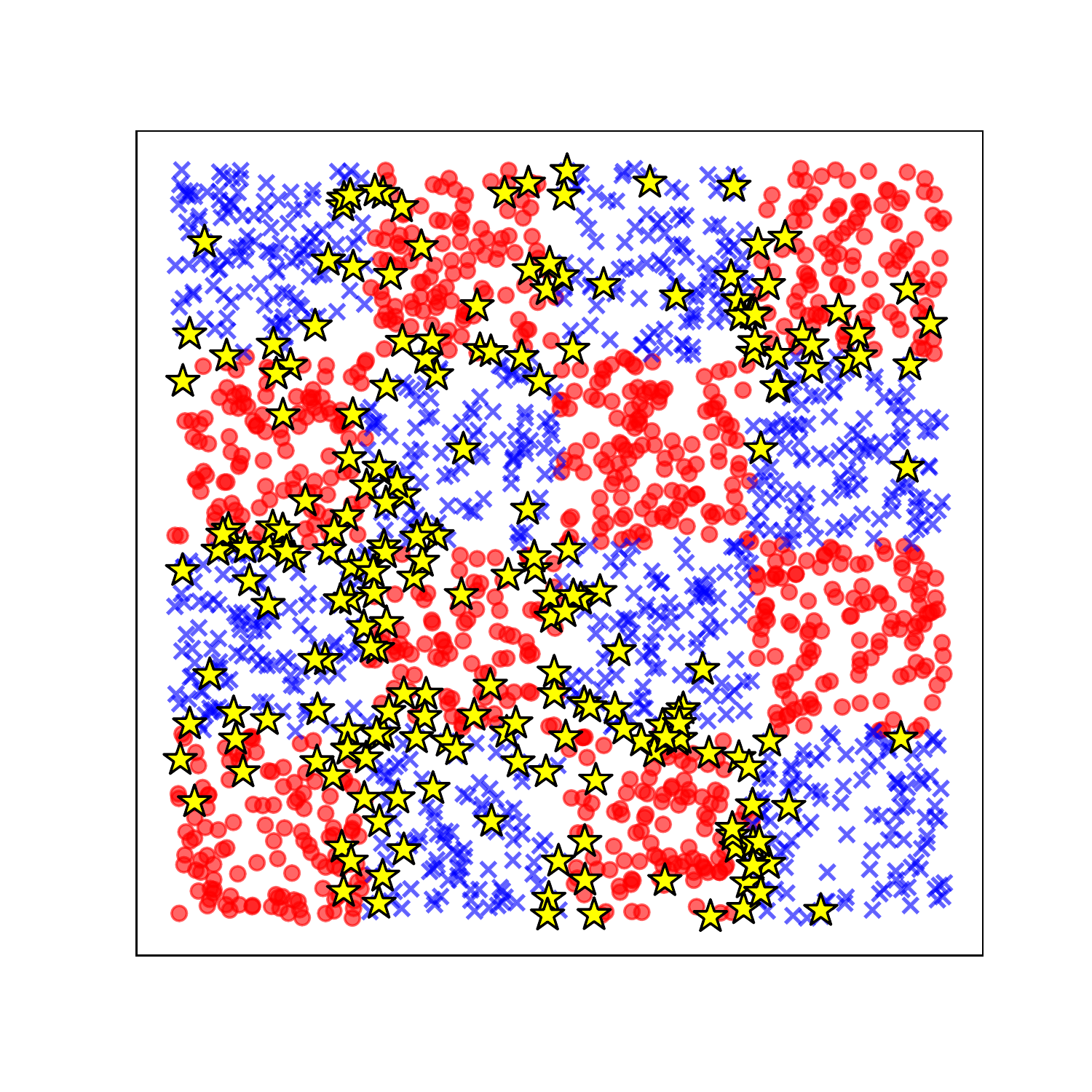}  
      \vspace{-0.4in}
      \caption{{Probit-Uncertainty}}
      \label{fig:probit-unc-choices}
    \end{subfigure}
    \caption{Acquisition function choices on the Checkerboard dataset. Yellow stars show the $200$ points chosen by each of the given acquisition functions.\vspace{-0in}}
    \label{fig:checkerboard-choices}
\end{figure}
The Checkerboard dataset 
consists of $2,000$ points uniformly sampled on the unit square $[0,1]^2 \subset \mbb R^2$, and we divide into two classes based on a $4 \times 4$ checkerboard pattern. For each of the five trials, we choose ten points uniformly at random to label initially (five from each class), and then sequentially choose $200$ query points via our list of acquisition functions. 
Similar to~\citep{kushnir_diffusion-based_2020}, we showcase this dataset because successful active learning in this dataset requires properly ``exploring'' the many different clusters as well as ``exploiting'' the learned decision boundaries efficiently. The best performing methods are the {\bf MC} methods in the {\bf GR} and {\bf Probit} models, as well as {\bf Probit-MBR}. These methods not only identify each of the clusters in the grid (Fig.~\ref{fig:gr-mc-choices},~\ref{fig:probit-mc-choices}) but also explore the decision boundaries between clusters. In the {\bf Probit-Uncertainty}(Fig.~\ref{fig:probit-unc-choices}) and {\bf HF-MBR}(Fig.~\ref{fig:hf-mbr-choices}), the methods have not explored the extent of the clustering structure and do not reach as high of accuracy (Fig.~\ref{fig:checkerboard}). Conversely, the {\bf VOpt} acquisition function in each model only identifies points that are representative of each of the clusters. As seen in Figs.~\ref{fig:hf-vopt-choices} and~\ref{fig:gr-vopt-choices}, these acquisition functions have not explored the boundaries between the clusters and so do not achieve as high of accuracy.

%
%
\vspace{-0.0in}
\subsection{MNIST}
\begin{figure}
    \begin{subfigure}{.32\textwidth}
      \centering
      \includegraphics[width=\linewidth]{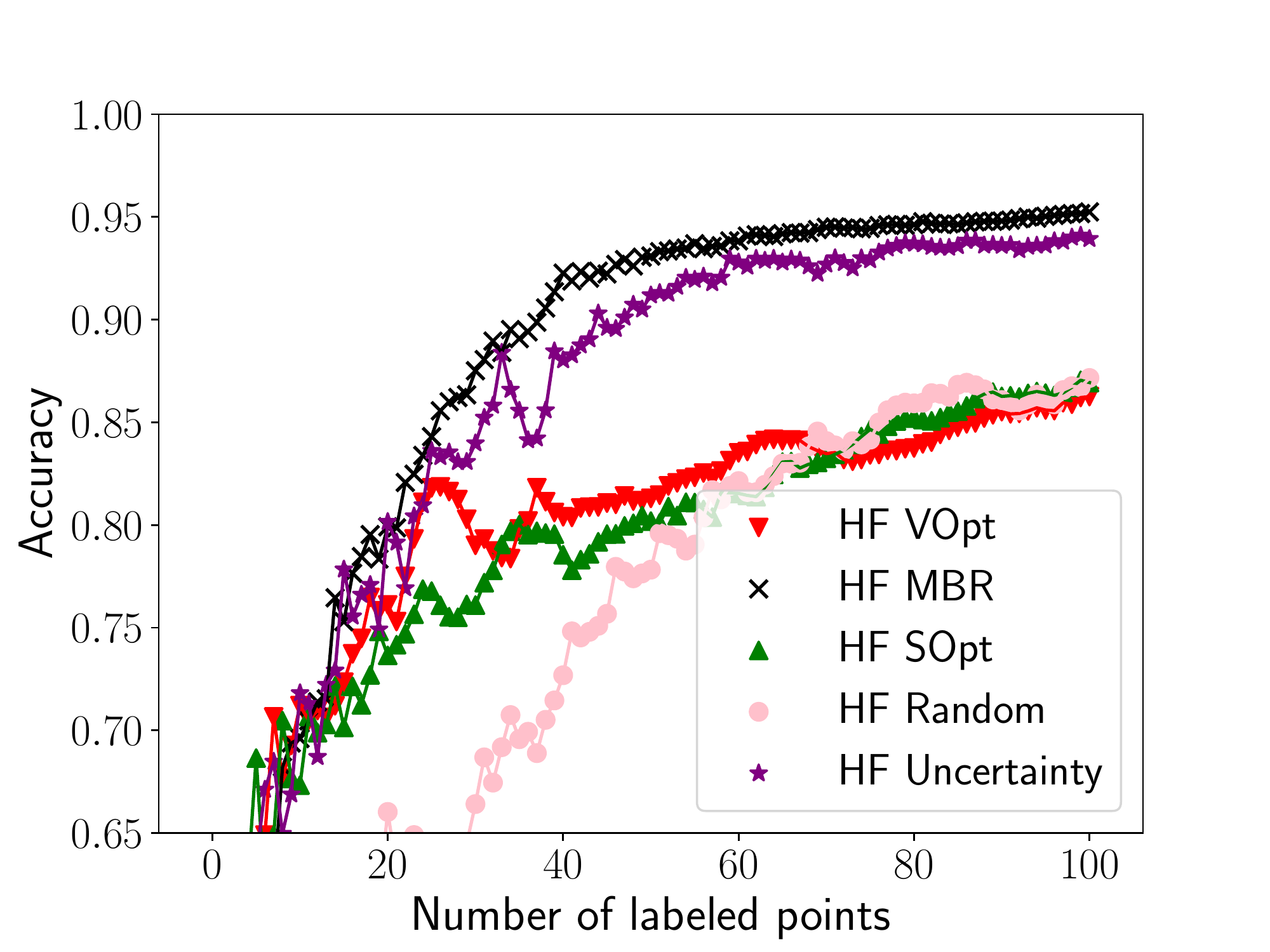}  
      \caption{HF }
      \label{fig:mnist-hf}
    \end{subfigure}
    \begin{subfigure}{.32\textwidth}
      \centering
      \includegraphics[width=\linewidth]{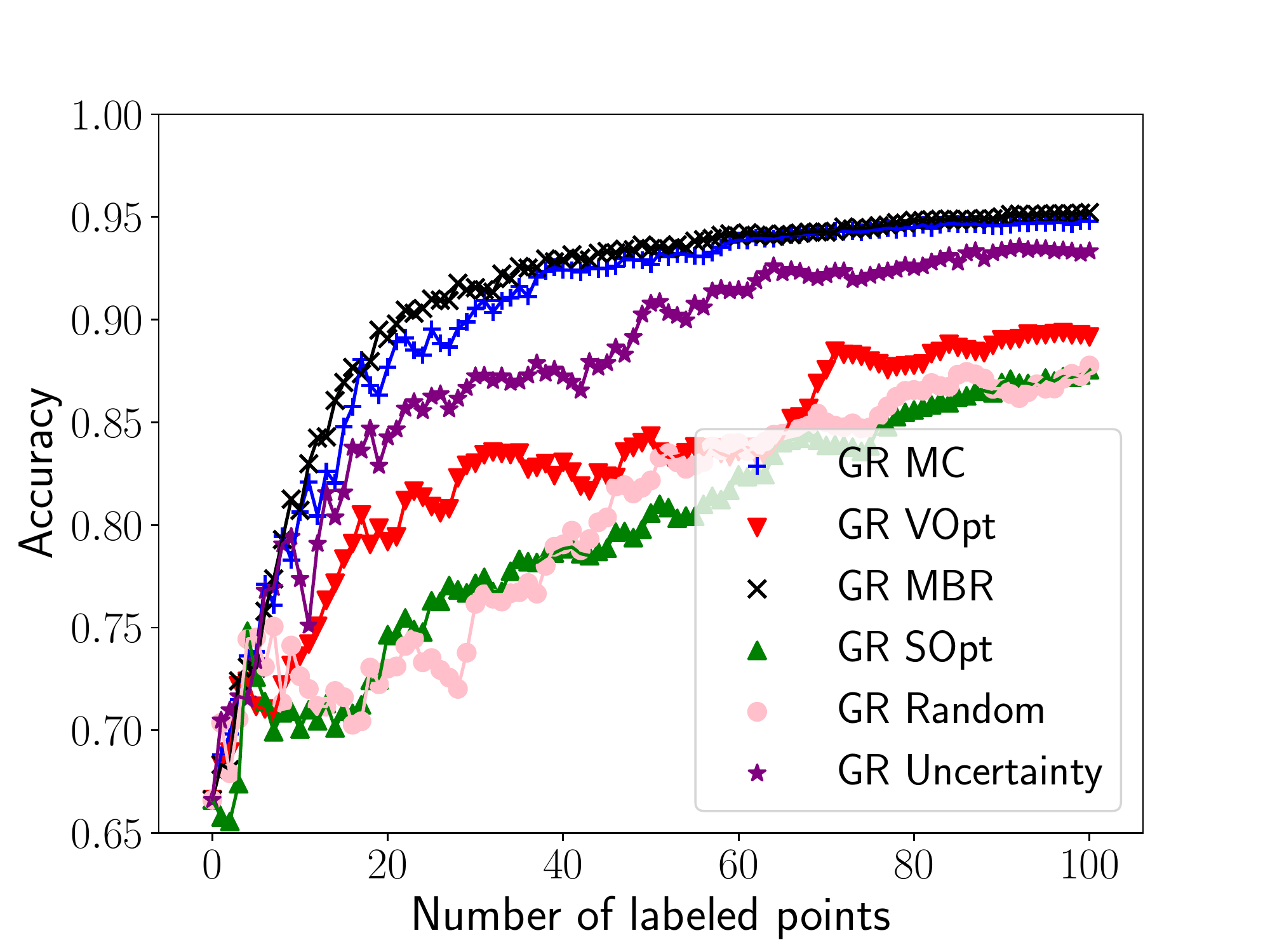}  
      \caption{GR}
      \label{fig:mnist-gr}
    \end{subfigure}
    \begin{subfigure}{.32\textwidth}
      \centering
      \includegraphics[width=\linewidth]{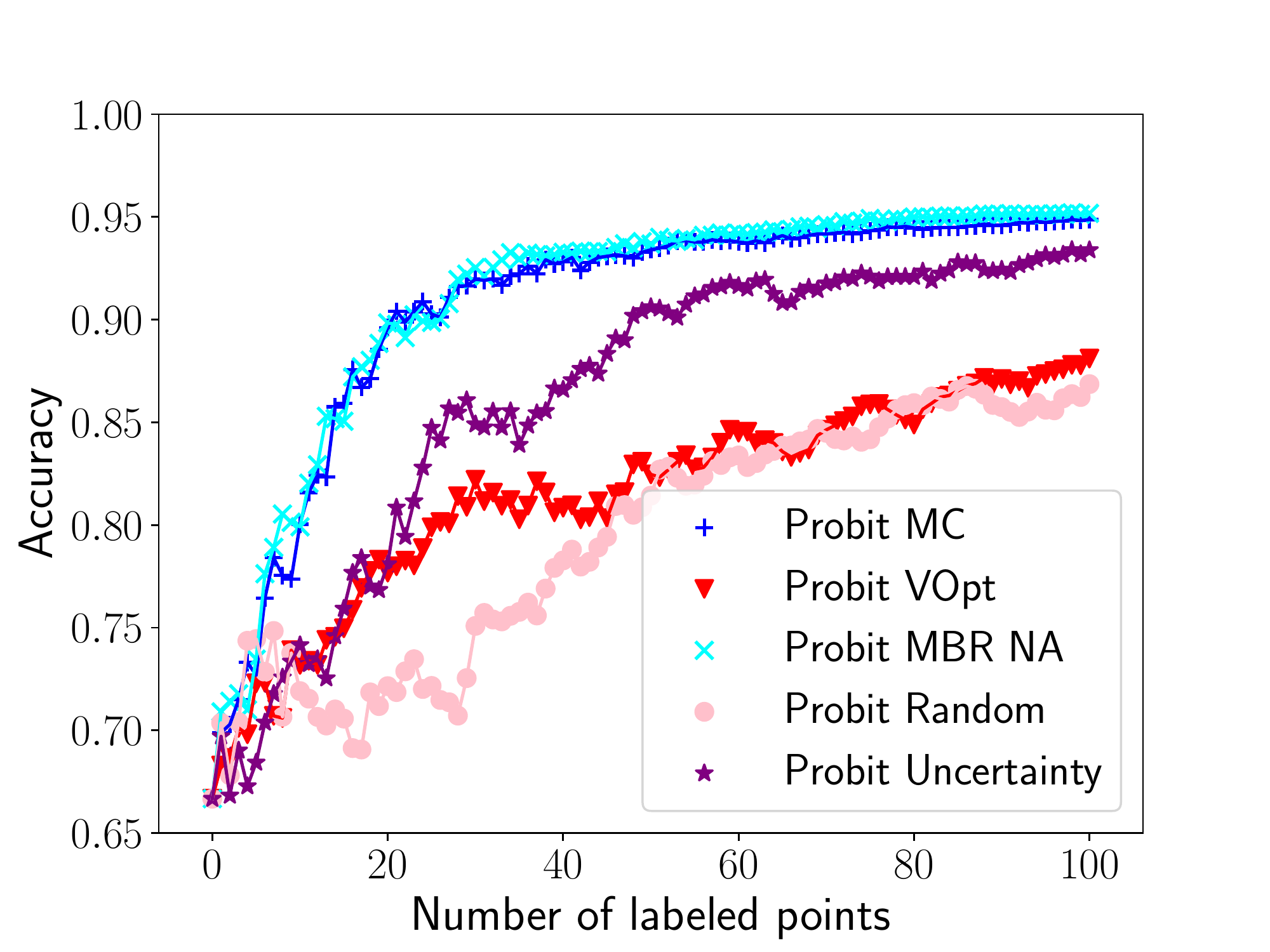}  
      \caption{Probit}
      \label{fig:mnist-p2}
    \end{subfigure}
    \caption{MNIST Dataset Results}
    \label{fig:mnist}
\end{figure}

MNIST \citep{lecun_mnist_1998} is a data set of 70,000 grayscale $28\times28$ pixel images of
handwritten digits (0-9). Each image is represented by a 784-dimensional vector $\bx_i$ and we normalize the pixel values to range from 0 to 1. We form a set of 4,000 data points by choosing uniformly at random 400 images from each digit. We construct a $15$-nearest neighbor graph among the data points with weights $w_{ij} = \exp(-\|\bx_i-\bx_j\|_2^2/380^2)$. We consider the binary classification problem of classifying even digits versus odd digits. For each of the five trials, we start with ten initial training points evenly distributed between the two classes (not necessarily among the digits) and use the active learners to query 100 points. The average classification accuracies are presented in Fig.~\ref{fig:mnist}.
Though the {\bf MBR} methods perform the best, they are more costly to compute than our competitive {\bf MC} acquisition functions.

\section{Conclusion and Future Directions}
Under this unifying Bayesian perspective of active learning in graph-based SSL, we use Laplace and Newton approximations to allow non-Gaussian models to employ acquisition functions previously only used in Gaussian models. We introduce a novel MC acquisition function that is both efficient to compute and provides competitive results. Future work could extend these results to batch-mode active learning, multi-class classification, and kernel methods other than graph-based SSL.


\acks{KM is supported by the DOD's National Defense Science and Engineering Graduate (NDSEG) Fellowship, while HL
and AB are supported by DARPA (grant FA8750-18-2-0066).}

\vskip 0.2in
\bibliography{sample}

\end{document}